\documentclass[11pt]{article}

\usepackage[final]{acl}

\usepackage{times}
\usepackage{latexsym}
\usepackage{tikz-dependency}

\usepackage[T1]{fontenc}
\usepackage{kotex}

\usepackage[utf8]{inputenc}

\usepackage{microtype}

\usepackage{inconsolata}

\usepackage{graphicx}

\title{Parser agreement and disagreement in L2 Korean UD: \\ Implications for human-in-the-loop annotation}

\author{
  Hakyung Sung\textsuperscript{1}\quad
  Gyu-Ho Shin\textsuperscript{2} \\
  \textsuperscript{1}Psychology, Rochester Institute of Technology\\
  \textsuperscript{2}Linguistics, University of Illinois Chicago\\
  \texttt{hksgla@rit.edu}\quad
  \texttt{ghshin@uic.edu}
}

\begin{document}
\maketitle
\begin{abstract}
We propose a simplified human-in-the-loop workflow for second language (L2) Korean morphosyntactic annotation by leveraging agreement between two domain-adapted parsers. We first evaluate whether parser agreement can serve as a proxy for annotation correctness by comparing it with independent human judgments. The results show strong correspondence between parser and human judgments, supporting the feasibility of semi-automatic L2-Korean UD annotation. Further analysis demonstrates that parser disagreements cluster in linguistically predictable domains such as grammatical-relation distinctions and clause-boundary ambiguity. While many disagreement cases are tractable for iterative model refinement, others reflect deeper representational challenges inherent in parsing and tagging L2-Korean corpora.
\end{abstract}

\section{Introduction}

Second language (L2) learner corpora consist of language samples produced by individuals acquiring an L2. Natural language processing (NLP) provides computational methods for extracting linguistic features (e.g., part-of-speech tags, grammatical relations). Together, these fields share the goal of empirically modeling L2 production at scale \cite{meurers2015learner}. However, applying NLP tools to L2 corpora has traditionally been considered problematic for two reasons. First, general-purpose models trained on well-edited native/first-language corpora are assumed to perform poorly on non-canonical learner language \cite{plank2016non}. Second, this assumption has been difficult to verify due to the lack of learner-language benchmarks.

In response, the Universal Dependencies (UD) framework has emerged as a principled foundation for annotating morphosyntactic features in L2 corpora \cite{masciolini2025annotating, zeman2025corpus}. UD offers cross-linguistic comparability while maintaining relative simplicity in morphosyntactic annotation and parsing \cite{deMarneffe2021UD}.

As UD-annotated learner corpora have become available, subsequent work has evaluated morphosyntactic models on such data, yielding mixed results. In high-resource languages such as L2 English, UD-based transformer models achieve over 90\% F1 in part-of-speech (POS) tagging and dependency parsing \cite{kyle2024evaluating}. In relatively under-resourced languages such as L2 Korean, performance varies by layer: morpheme-level tagging reaches F1 ≈ 88\%, while dependency parsing remains substantially lower (LAS ≈ 57\%) \cite{sung2025towards}. Performance also varies by genre and proficiency. For instance, written learner data yield lower parsing accuracy than spoken data in L2 English \cite{kyle2024evaluating}. In L2 Korean, higher-proficiency learners tend to produce longer and more syntactically complex sentences, which increase parsing difficulty; accordingly, proficiency is negatively correlated with dependency head accuracy ($r = -0.26$) \cite{sung2025towards}. Importantly, both studies report improved performance after fine-tuning on L2-annotated data, suggesting that domain adaptation partially mitigates these limitations.

Taken together, for researchers seeking to leverage NLP-based annotations in L2 research, the question is increasingly no longer whether morphosyntactic models can be applied to learner corpora, but how they can be effectively integrated into annotation workflows. Because performance varies across languages, proficiency levels, and genres, fully automatic annotation may remain insufficient. However, automatic pre-annotation paired with human verification offers a practical alternative. In this context, the present study explores a simplified human-in-the-loop workflow for L2-Korean UD annotation, testing whether parser agreement can guide selective human review without compromising annotation reliability.

\raggedbottom

\section{Related work}

\subsection{Use of morphosyntactic annotation in L2 learner corpora research}
\label{sec2:1}
\paragraph{Part-of-Speech tags:} Part-of-Speech (POS) tags have been widely employed in learner corpora research. In L2 English, for example, phraseological competence (i.e., the use of semi-/prefabricated expressions) has been examined through the automatic extraction of patterns from POS-tagged corpora (e.g., \citealp{granger2014use}). POS tags have also been used to measure lexical richness and disambiguate homographs in L2 Spanish \cite{diez2024measuring}. In the case of L2 Korean, language-specific POS tagsets enabled the representation of fine-grained morphemic distinctions within words \cite{sung2024empirical}.

\paragraph{Dependency relations:} Syntactic information derived from dependency relations has likewise supported diverse corpus-based investigations. For instance, dependency-based phraseological units have been analyzed in L2 Dutch to examine the lexis–grammar interface \cite{rubin2025exploring}, while dependency representations have been used to assess lexical and syntactic complexity in L2 Russian \cite{kisselev2022measuring}. In L2 English, prior work examined n-grams within specific dependency relations (e.g., \citealp{paquot2019phraseological}), verb–argument constructions and related predicate–argument patterns (e.g., \citealp{kyle2017assessing}), and broader measures of syntactic complexity (e.g., \citealp{kyle2018measuring}). Similarly, \citet{hao2024syntactic} employed dependency parsing to investigate syntactic complexity in L2-Chinese writing.

\begin{table*}[htbp]
\centering
\resizebox{\linewidth}{!}{
\begin{tabular}{llll}
\hline
\textbf{Language (domain)} & \textbf{Annotation method(s)}  & \textbf{Reference} \\
\hline
Chinese (written) & Manual & \citet{lee2017towards} \\
English (written) & Manual & \citet{berzak2016universal} \\
English (spoken) & Manual & \citet{kyle2022dependency} \\
Italian (written) & Semi-automatic & \citet{di2019towards} \\
Korean (written) & Manual; Semi-automatic & \citet{sung2024constructing, sung2025ud} \\
Russian (written) & Manual (single annotator) & \citet{rozovskaya2024universal} \\
Spanish (written) & Manual & \citet{pulido2025speak} \\
Swedish (written) & Semi-automatic & \citet{swell_with_pride} \\
\hline
\end{tabular}}
\caption{Overview of UD annotation practices in L2 learner corpora}
\label{tab:1}
\end{table*}

\subsection{Reliability of morphosyntactic annotation on L2 corpora}

Although previous studies (as exemplified in Section~\ref{sec2:1}) have reported important empirical findings based on extracted morphosyntactic features, their validity depends in part on annotation reliability. If automatic analyses are inaccurate, resulting conclusions may be compromised. While several studies evaluated the performance of NLP models on L2 corpora (e.g., \citealp{berzak2016universal}), findings have been mixed and often limited in scope. As noted by \citet{kyle2024evaluating}, many investigations have focused on isolated components (e.g., selected POS tags or dependency relations) rather than overall morphosyntactic performance. Moreover, earlier studies relied on neural architectures trained primarily on well-edited standard-language data (e.g., news articles), without adaptation to L2 learner language.

A recent advance has been the adoption of domain adaptation techniques, in which annotated L2 treebanks are incorporated into model training to improve annotation quality \cite{kyle2024evaluating, sung2025towards}. Although effective, this approach presupposes the availability of reliable L2 annotations. This aspect makes it important to examine how such annotations are produced in existing L2 corpora. 

\subsection{UD annotation practices in L2 corpora}

Over the past decade, an increasing number of L2 corpora have been annotated within the UD framework for diverse research purposes \cite{zeman2025corpus}. Our review identified eight such corpora to date. \citet{masciolini2025annotating} provide a detailed comparison, outlining their design characteristics (e.g., modality, size, annotation status) and the strategies adopted to address L2-specific phenomena, including ill-formed or non-canonical constructions.

Here, we examine the annotation methodologies underlying these corpora, focusing on whether morphosyntactic annotation was conducted either fully manually or through semi-automatic procedures (i.e., automatically annotated and then corrected by humans; see Table~\ref{tab:1}). Most UD-based L2 corpora relied on fully manual annotation, with relatively few adopting automatic approaches supplemented by human correction. While manual annotation supports quality control (e.g., through inter-annotator reliability), it is resource-intensive, difficult to scale, and challenging to replicate consistently across projects and annotator teams. Hybrid approaches that combine automatic processing with human oversight may therefore provide a more efficient and reproducible alternative.

\subsection{Human-in-the-Loop annotation via model agreement}

Human-in-the-Loop (HITL) machine learning broadly refers to workflows in which human expertise is intentionally integrated into automated systems to guide, validate, or correct model behavior \cite{mosqueira2023human}. Rather than replacing automation, such approaches strategically combine machine efficiency with human judgment, enhancing scalability while preserving reliability. They have been increasingly adopted in domains where full automation is unreliable or where high-stakes decisions require human oversight \cite{amershi2014power, holzinger2016interactive}.

Within the broader HITL taxonomy \cite{holmberg20}, active learning represents a prominent paradigm in which models select informative or uncertain instances for human annotation \cite{settles2009active}. Here, annotators function as oracles, and their feedback is used to iteratively refine the model. Active learning has been widely applied in NLP tasks such as POS tagging, dependency parsing, and text classification to reduce annotation cost while maintaining performance \cite{ringger2008assessing}. 

Beyond uncertainty-based selection, prior work in dependency parsing has shown that simple ensemble strategies (i.e., agreement-based voting across multiple parsers) can produce robust predictions without requiring complex meta-modeling \cite{surdeanu2010ensemble}. This suggests that model agreement can serve as a complementary signal of confidence. Building on this idea, the present study leverages parser agreement to guide selective human intervention, using disagreement cases as candidates for targeted review.

\section{Experiment}

In this exploratory study, we examine a simplified HITL workflow for L2-Korean UD annotation. Specifically, we compare the outputs of two independently fine-tuned parsers, treating agreement as a proxy for reliable annotation and disagreement as a signal for targeted human review. We assess whether such a setup can support more efficient annotation in future L2-Korean corpora. The study addresses the following research questions:

\begin{enumerate}
    \item Can parser agreement reliably serve as a proxy for human annotation agreement?

    \item How much manual correction is required to resolve parser disagreements?
    
    \item Which morphosyntactic categories exhibit disagreement, and how can these patterns inform annotation refinement?
\end{enumerate}

\subsection{Proposed framework}
\label{sec:3.1}

The proposed annotation framework consists of three steps.

\paragraph{Step 1: Automatic annotation.}
Two domain-adapted parsers—\textit{Stanza} \cite{qi2020stanza} and \textit{Trankit} \cite{van2021trankit}—were applied.\footnote{Stanza is a neural pipeline that performs joint tokenization, POS tagging, lemmatization, and dependency parsing using BiLSTM-based and transition-based components, while Trankit is a transformer-based multilingual pipeline built on XLM-R representations for joint morphosyntactic analysis. Both models were fine-tuned on the UD-KSL training set \cite{sung2025second}, a learner corpus of L2 Korean writing annotated with morpheme-level segmentation, XPOS tags, and dependencies.} We examined four layers: \texttt{LEMMA}, \texttt{XPOS}, \texttt{HEAD}, and \texttt{DEPREL}.\footnote{\texttt{UPOS} was excluded because it is deterministically derived from \texttt{XPOS} under the current annotation scheme \citep{sung2025ud}.} Table~\ref{tab:2} reports in-domain performance on the UD-KSL test set \cite{sung2025second}. 

\begin{table}[t]
\centering
\begin{tabular}{lcc}
\hline
Metric & Stanza & Trankit \\
\hline
\texttt{LEMMA} & 95.64 & 88.84 \\
\texttt{XPOS}  & 89.72 & 91.81 \\
\texttt{UAS}   & 85.53 & 92.28 \\
\texttt{LAS}   & 80.36 & 89.13 \\
\hline
\end{tabular}
\caption{Performance comparison (F1 scores) of fine-tuned Stanza and Trankit models on the L2K-UD test dataset}
\label{tab:2}
\end{table}

\textbf{Step 2: Cross-model comparison.}
Parser outputs were compared at the token level. Tokens with identical outputs were provisionally accepted, whereas any disagreement triggered human review.

\textbf{Step 3: Human adjudication.}
Two trained annotators independently reviewed the flagged tokens. If they assigned identical annotations at the token level, their decision was adopted as the gold label. In cases of disagreement, a third annotator (one of the authors) adjudicated by reviewing both model outputs and the independent annotations to assign the final label.

\subsection{Evaluation of parser agreement} 

Given the exploratory nature of this study, we first collected fully independent annotations from both human annotators for all sentences before restricting review to parser-disagreement cases. This design allowed us to assess whether parser agreement could serve as a proxy for human annotation agreement. Specifically, we measured how often parser agreement coincided with inter-annotator agreement. Under the assumption that alignment is sufficiently strong (i.e., human agreement exceeds 90\% within parser-agreement cases), parser-agreement cases could be retained automatically in future annotation rounds, enabling human effort to focus primarily on disagreement cases.

\subsection{Pipeline validation}

Prior to full-scale annotation of the target corpus, we conducted a small-scale validation experiment to evaluate whether the proposed semi-automatic pipeline introduced annotation noise. We randomly sampled 500 L2-Korean sentences from the KoLLA corpus \cite{lee2009annotation}, none of which had been used to train the fine-tuned models. Annotation proceeded incrementally. In each round, a batch of 100 sentences was annotated using the framework described in Section~\ref{sec:3.1}, with human review limited to tokens where the two models disagreed. Adjudicated annotations were then incorporated into the training data, and both models were fine-tuned for 10 epochs on the expanded dataset. After each round, performance was evaluated on the UD-KSL test set to determine whether accuracy improved, stabilized, or declined. This procedure was repeated for five rounds.

\begin{figure}[h!]
\centering
\includegraphics[width=\linewidth]{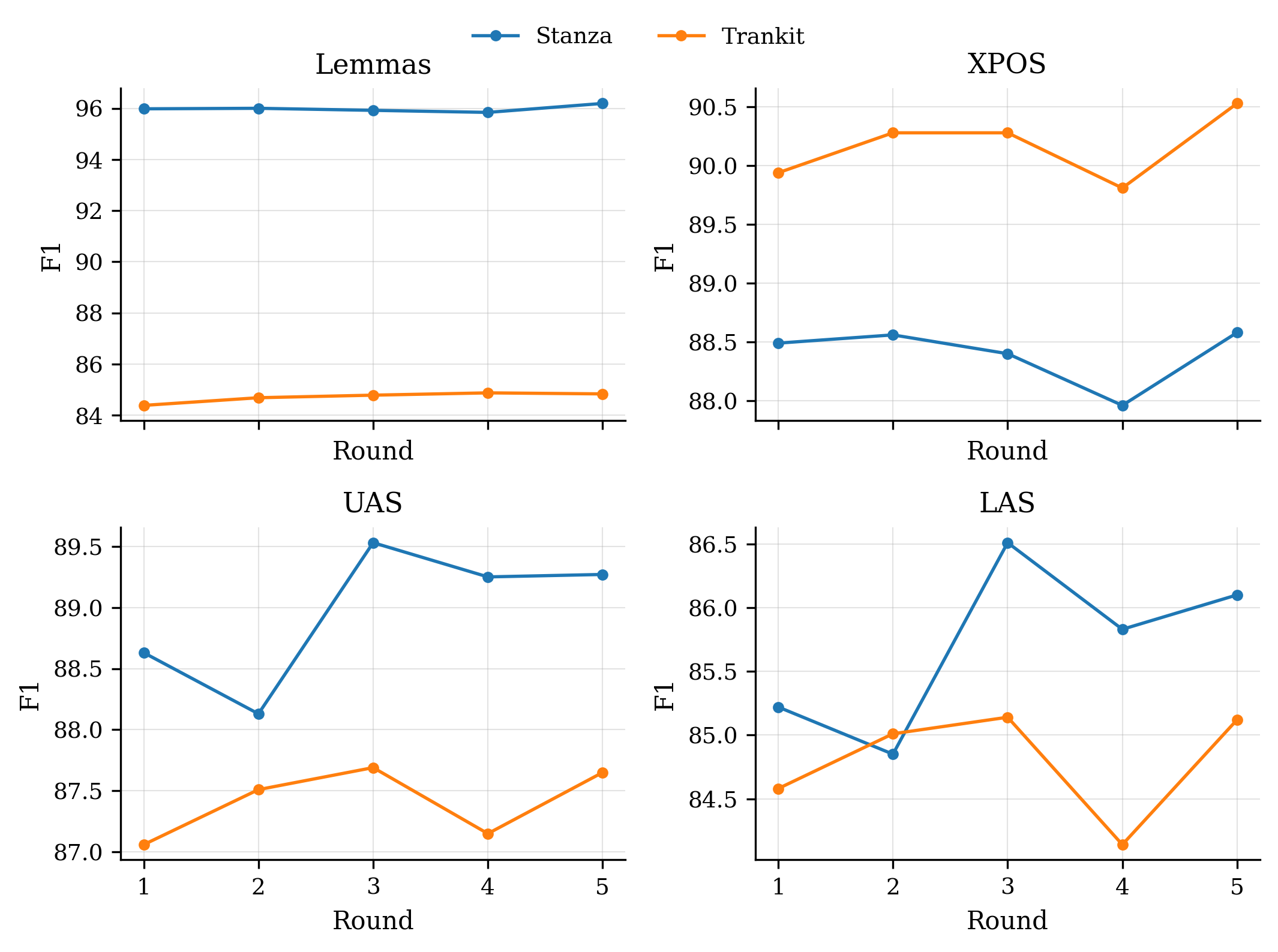}
\caption{Model performance on the test set across fine-tuning rounds}
\label{fig:1}
\end{figure}

Figure~\ref{fig:1} presents model performance on the test set across incremental fine-tuning rounds. Overall, performance remained stable: both Stanza and Trankit maintained accuracies of approximately 85\% or higher, with no observable decline. These results indicate that the proposed pipeline did not introduce substantial degradation in annotation quality during incremental fine-tuning.

\section{Results}

\subsection{Annotations}

Two trained annotators annotated a total of 2,208 sentences drawn from argumentative essays written by Japanese and English learners of Korean over a one-month period. We adopted the morphosyntactic annotation scheme developed in prior L2 Korean UD annotation work, most recently detailed in \citet{sung2025ud}.

\subsection{Parser agreement as a proxy for human annotation agreement}

To evaluate whether parser agreement can serve as a proxy for human annotation agreement, we calculated token-level agreement rates, excluding punctuation. Table~\ref{tab:3} summarizes the results. Across all features, the two parsers agreed on 82\% of token-level decisions. Within these parser-agreement cases, the human annotators also agreed on 93\% of instances, with agreement exceeding 90\% for all four features. These findings indicate that parser agreement closely corresponded to independent human agreement, supporting its use as a practical proxy for human annotation agreement in the HITL workflow.

\begin{table}[h]
\centering
\resizebox{\linewidth}{!}{
\begin{tabular}{lcc}
\hline
\textbf{Feature} & \textbf{Model agreement} & \textbf{Human agreement} \\
\hline
\texttt{LEMMA}  & 78.09 & 96.97 \\
\texttt{XPOS}   & 84.83 & 91.97 \\
\texttt{HEAD}   & 78.62 & 90.22 \\
\texttt{DEPREL} & 88.78 & 92.55 \\
\hline
Average & 82.58 & 92.93 \\
\hline
\end{tabular}}
\caption{Token-level agreement between the two parsers and corresponding human annotator agreement rates (punctuation excluded)}
\label{tab:3}
\end{table}

\subsection{Human intervention following model mismatch}

Based on the agreement results, we next evaluated a workflow in which human annotators intervened only when the two models disagreed.\footnote{Prior to analysis, tokenization mismatches were resolved to ensure proper token-level alignment. All reported agreement and intervention rates are based on the aligned data.} At the token level, 7,994 out of 25,814 tokens (31\%) required human correction in at least one morphosyntactic feature.\footnote{Because a single token may require correction in multiple features, counts are not mutually exclusive.} 

Feature-level correction counts are provided in Table~\ref{tab:4}, and feature-level adjudication corrections are summarized in Table~\ref{tab:5}. A total of 2,019 out of 25,814 tokens (8\%) required further modification after initial review. These cases reflect instances in which the two annotators did not converge, necessitating adjudication by a third annotator. 

\begin{table}[h]
\centering
\begin{tabular}{lccc}
\hline
\textbf{Feature} & \textbf{Human} & \textbf{Total} & \textbf{Rate (\%)} \\
\hline
\texttt{LEMMA} & 4,485 & 25,814 & 17.37 \\
\texttt{XPOS}    & 2,263 & 25,814 & 8.77  \\
\texttt{HEAD}   & 3,798 & 25,814 & 14.71 \\
\texttt{DEPREL} & 1,713 & 25,814 & 6.64  \\
\hline
\end{tabular}
\caption{Feature-level human corrections following model mismatches}
\label{tab:4}
\end{table}

\begin{table}[h]
\centering
\begin{tabular}{lccc}
\hline
\textbf{Feature} & \textbf{Fixed} & \textbf{Total} & \textbf{Rate (\%)} \\
\hline
\texttt{LEMMA}  & 1,125 & 25,814 & 4.36 \\
\texttt{XPOS}   & 1,461 & 25,814 & 5.66 \\
\texttt{HEAD}   & 581   & 25,814 & 2.25 \\
\texttt{DEPREL} & 1,153 & 25,814 & 4.47 \\
\hline
\end{tabular}
\caption{Feature-level third-annotator corrections}
\label{tab:5}
\end{table}

Overall, these findings demonstrate how parser disagreement can structure a tiered annotation workflow. Restricting human review to model-disagreement cases substantially reduces effort, with nearly 70\% of tokens requiring no intervention after alignment. These tokens likely represent morphosyntactic categories that are relatively stable and well captured by the fine-tuned models. Most remaining disagreement cases were resolved through agreement between two annotators, suggesting that they are tractable.

In contrast, the 8\% of tokens requiring third-annotator adjudication represent persistent disagreement across both models and trained annotators. These instances often involved structurally complex or potentially ambiguous linguistic units. Although tentative, such residual disagreement at high overall accuracy may reflect not only model limitations but also indeterminacy in linguistic categories or variation in annotation conventions \cite{manning2011part}. 

\subsection{Disagreement analysis}

To further characterize these disagreement patterns, we analyzed where and why the models diverged.

\subsubsection{Distribution of dependency-relation disagreements}

We first conducted a focused analysis of disagreement patterns in the dependency-relation (\texttt{DEPREL}) layer, as dependency-relation labeling exhibited relatively high disagreement rates. Disagreements were classified according to the primary syntactic decision involved: (1) grammatical relation identification, (2) clause-boundary and clause-type differentiation, (3) discourse-level structural organization, and (4) modifier attachment.

\begin{itemize}
    \item \textbf{Grammatical relation identification}: This category captures instability in assigning core grammatical relations (e.g., subject, object, oblique), including contrasts such as \texttt{nsubj–obj}, \texttt{nsubj–obl}, and \texttt{obj–obl}.
    
    \item \textbf{Clause boundary}: This category reflects uncertainty in clause typing and hierarchical embedding, including distinctions among adjectival (relative), adverbial, and complement clauses (e.g.,  \texttt{acl–advcl}, \texttt{advcl–ccomp}, and \texttt{advcl–root}).
    
    \item \textbf{Discourse-level organization}: This category involves higher-level decisions at the syntax–discourse interface, such as coordination scope, root status, and left dislocation (e.g., topic-marked elements). Recurrent contrasts included \texttt{dislocated–nsubj}, \texttt{root–conj}, and \texttt{conj–advcl}.

    \item \textbf{Modifier attachment}: This category captures ambiguity in determining the structural status or attachment site of modifiers, with contrasts such as \texttt{amod–acl} and \texttt{nmod–obl}.
\end{itemize}

Table~\ref{tab:6} summarizes mismatch frequencies across categories. To illustrate these patterns more concretely, representative examples from the annotated texts are provided below.\footnote{Sentences have been streamlined for clarity.}

\begin{table}[t]
\centering
\begin{tabular}{l r}
\hline
\textbf{Mismatch type} & \textbf{Count} \\
\hline
Grammatical relation & 263 \\
Clause boundary & 206 \\
Discourse / structure & 235 \\
Modifier attachment & 152 \\
\hline
\end{tabular}
\caption{Major categories of dependency-relation disagreements between the two parsers. Counts indicate the total number of mismatches in each category.}
\label{tab:6}
\end{table}

\begin{figure}[h!]
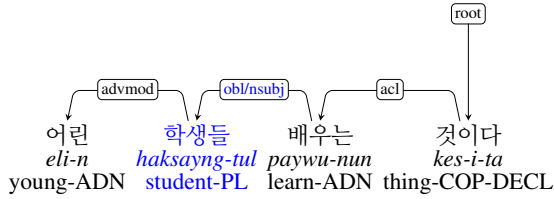

    \centering
    \resizebox{\linewidth}{!}{
        \begin{dependency}
            \begin{deptext}
            어린 \& \textcolor{blue}{학생들} \& 배우는 \& 것이다\\
            \textit{eli-n} \& \textcolor{blue}{\textit{haksayng-tul}} \& \textit{paywu-nun} \& \textit{kes-i-ta} \\
            young-ADN \& \textcolor{blue}{student-PL} \& learn-ADN \& thing-COP-DECL \\
            \end{deptext}
            \depedge{2}{1}{advmod} 
            \depedge{3}{2}{\textcolor{blue}{obl/nsubj}}
            \depedge{4}{3}{acl}
            \deproot{4}{root}
        \end{dependency}
    }
    \caption{Grammatical-relation ambiguity under case marker omission 1. The nominal 학생들 ‘students’ was tagged as either \texttt{obl} or \texttt{nsubj} across the parsers; contextual interpretation favors \texttt{nsubj}. \footnotesize{(Translated as ‘(It) is that young students learn.’)}}
    \label{fig:2}
\end{figure}

\begin{figure}[h!]
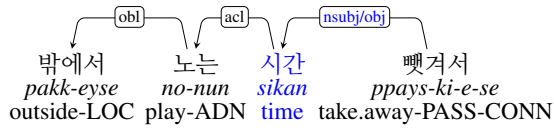

    \centering
    \resizebox{\linewidth}{!}{
        \begin{dependency}
            \begin{deptext}
            밖에서 \& 노는 \& \textcolor{blue}{시간} \& 뺏겨서 \\
            \textit{pakk-eyse} \& \textit{no-nun} \& \textcolor{blue}{\textit{sikan}} \& \textit{ppays-ki-e-se} \\
            outside-LOC \& play-ADN \& \textcolor{blue}{time} \& take.away-PASS-CONN \\
            \end{deptext}
            \depedge{2}{1}{obl} 
            \depedge{3}{2}{acl} 
            \depedge{4}{3}{\textcolor{blue}{nsubj/obj}}
            
        \end{dependency}
    }
    \caption{Grammatical-relation ambiguity under case marker omission 2. The nominal 시간 ‘time’ received conflicting tags (\texttt{nsubj} vs.\ \texttt{obj}); contextual interpretation favors \texttt{obj}. \footnotesize{(Translated as ‘(When) time spent playing outside is taken away.’)}}
    \label{fig:3}
\end{figure}

Grammatical-relation ambiguity (e.g., \texttt{nsubj--obj}, \texttt{nsubj--obl}) accounts for the largest share of mismatches. These alternations typically arise when case-marking is underspecified or omitted, obscuring whether a nominal is analyzed as subject, object, or oblique. For example, when nominative (Figure~\ref{fig:2}) or accusative case markers (Figure~\ref{fig:3}) are dropped, a preverbal noun phrase may be ambiguously interpreted, leading to divergent grammatical relation identifications across parsers.

\begin{figure}[!]
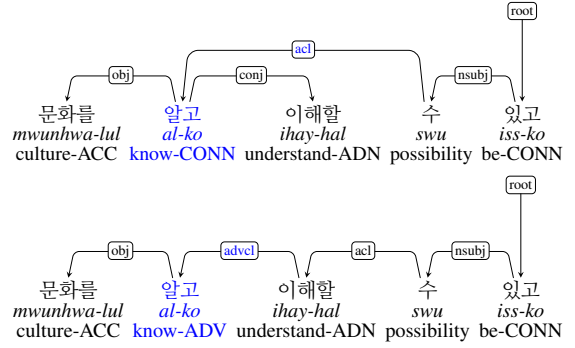

    \centering
    
    \resizebox{\linewidth}{!}{
        \begin{dependency}
            \begin{deptext}
            문화를 \& \textcolor{blue}{알고} \& 이해할 \& 수 \& 있고\\
            \textit{mwunhwa-lul} \& \textcolor{blue}{\textit{al-ko}} \& \textit{ihay-hal} \& \textit{swu} \& \textit{iss-ko}\\
            culture-ACC \& \textcolor{blue}{know-CONN}  \& understand-ADN \& possibility \& be-CONN \\
            \end{deptext}
            \depedge{2}{1}{obj} 
            \depedge{4}{2}{\textcolor{blue}{acl}}
            \depedge{2}{3}{conj}
            \depedge{5}{4}{nsubj}
            \deproot{5}{root}
        \end{dependency}
    }
    \vspace{0.8em}
    \resizebox{\linewidth}{!}{
        \begin{dependency}
            \begin{deptext}
            문화를 \& \textcolor{blue}{알고} \& 이해할 \& 수 \& 있고\\
            \textit{mwunhwa-lul} \& \textcolor{blue}{\textit{al-ko}} \& \textit{ihay-hal} \& \textit{swu} \& \textit{iss-ko}\\
            culture-ACC \& \textcolor{blue}{know-ADV} \& understand-ADN \& possibility \& be-CONN \\
            \end{deptext}
            \depedge{2}{1}{obj} 
            \depedge{3}{2}{\textcolor{blue}{advcl}}
            \depedge{4}{3}{acl}
            \depedge{5}{4}{nsubj}
            \deproot{5}{root}
        \end{dependency}
    }

\caption{Clause-boundary ambiguity 1. The connective verb 알고 (‘know-CONN/ADV’) is ambiguous. In the first analysis, the clause is treated as an adnominal modifier (\texttt{acl}); in the second, it is analyzed as an adverbial clause (\texttt{advcl}); the appropriate annotation cannot be determined from the sentence in isolation. \footnotesize{(Translated as Top: ‘[One] can know and understand the culture.’ Bottom: ‘After knowing the culture, [one] can understand it.’)}}
\label{fig:4}    
\end{figure}

Clause-boundary ambiguity (e.g., \texttt{acl–advcl}, \texttt{advcl–ccomp}) reflects uncertainty in clause typing and hierarchical embedding. Figure~\ref{fig:4} illustrates a case in which a clause can be analyzed either as an adnominal modifier (top) or as a subordinate clause within the predicate domain (bottom).

Compared to grammatical-relation ambiguity, clause-boundary ambiguity poses greater challenges for two reasons. First, the polyfunctionality of the connective 고 \textit{-ko} in Korean frequently triggers disagreement, as its interpretation depends on discourse-semantic cues rather than overt syntactic marking. Second, clause-typing uncertainty often interacts with higher-level structural mismatches (e.g., \texttt{conj–advcl}), corresponding to the third category of disagreement. Such cases therefore require adjudication informed by broader sentential or discourse context.

\begin{figure}[h!]
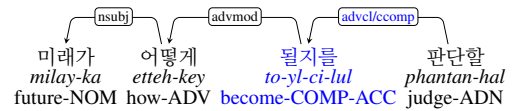

    \centering
    \resizebox{0.9\linewidth}{!}{
        \begin{dependency}
            \begin{deptext}
            미래가 \& 어떻게 \& \textcolor{blue}{될지를} \& 판단할 \\
            \textit{milay-ka} \& \textit{etteh-key} \& \textcolor{blue}{\textit{to-yl-ci-lul}} \& \textit{phantan-hal}\\
            future-NOM \& how-ADV \& \textcolor{blue}{become-COMP-ACC} \& judge-ADN \\
            \end{deptext}
            \depedge{2}{1}{nsubj} 
            \depedge{3}{2}{advmod}
            \depedge{4}{3}{\textcolor{blue}{advcl/ccomp}}
        \end{dependency}
    }
    \caption{Clause-type disagreement 2. The embedded clause 될지를 (‘how [it] will become’) functions as a clausal complement (\texttt{ccomp}) of the matrix predicate 판단할 (‘judge’), a pattern that one parser consistently failed to capture; morphosyntactic cues (i.e., the complementizer 지 and accusative marker 를) favor a \texttt{ccomp}.\footnotesize{(Translated as: ‘[One] can judge how the future will turn out.’)}}
\label{fig:5}    
\end{figure}

Meanwhile, not all clause-type disagreements arise from context-dependent ambiguity. In \texttt{advcl–ccomp} mismatches, some cases instead appear to reflect model difficulty in learning complement structures headed by the -지 (\textit{-ci}) complementizer (Figure~\ref{fig:5}).

Discourse-related ambiguity (e.g., \texttt{dislocated–nsubj}) reflects instability at the syntax-discourse interface, particularly in topic-prominent constructions where left-dislocated elements may be misanalyzed as canonical subjects.\footnote{Examples of structural mismatches (e.g., \texttt{root–conj}) are not presented separately, as they typically co-occur with clause-boundary ambiguities discussed above.} 

\begin{figure}[h!]
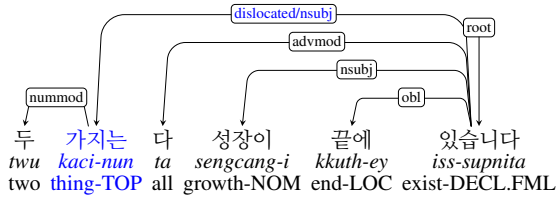

    \centering
    \resizebox{\linewidth}{!}{
        \begin{dependency}
            \begin{deptext}
            두 \& \textcolor{blue}{가지는} \& 다 \& 성장이 \& 끝에 \& 있습니다\\
            \textit{twu} \&  \textcolor{blue}{\textit{kaci-nun}} \& \textit{ta} \& \textit{sengcang-i} \& \textit{kkuth-ey} \& \textit{iss-supnita}\\
            two \& \textcolor{blue}{thing-TOP} \& all \& growth-NOM \& end-LOC \& exist-DECL.FML \\
            \end{deptext}
            \depedge{2}{1}{nummod} 
            \depedge{6}{2}{\textcolor{blue}{dislocated/nsubj}}
            \depedge{6}{3}{advmod}
            \depedge{6}{4}{nsubj}
            \depedge{6}{5}{obl}
            \deproot{6}{root}
        \end{dependency}
    }
    \caption{Discourse-level misanalysis. The topic-marked noun phrase 가지는 (‘things-TOP’) functions as \texttt{dislocated}, a pattern that one parser consistently failed to capture. \footnotesize{(Translated as ‘As for the two things, both ultimately result in growth.’)}}
\label{fig:6}    
\end{figure}

\begin{figure}[h!]
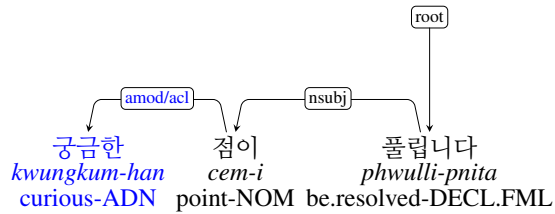

    \centering
    \resizebox{\linewidth}{!}{
        \begin{dependency}
            \begin{deptext}
            \textcolor{blue}{궁금한} \& 점이 \& 풀립니다\\
            \textcolor{blue}{\textit{kwungkum-han}} \& \textit{cem-i} \& \textit{phwulli-pnita}\\
            \textcolor{blue}{curious-ADN} \& point-NOM \& be.resolved-DECL.FML\\
            \end{deptext}
            \depedge{2}{1}{\textcolor{blue}{amod/acl}}
            \depedge{3}{2}{nsubj}
            \deproot{3}{root}
        \end{dependency}
    }
    \caption{Modifier attachment ambiguity 1. The form 궁금한 (‘curious-ADN’) can function either as a lexical adjectival modifier (\texttt{amod}) or as a reduced relative clause (\texttt{acl}) modifying \textit{cem} (‘point’). 
    \footnotesize{(Translated as ‘The questions are resolved.’)}}
\label{fig:7}    
\end{figure}

\begin{figure}[!]
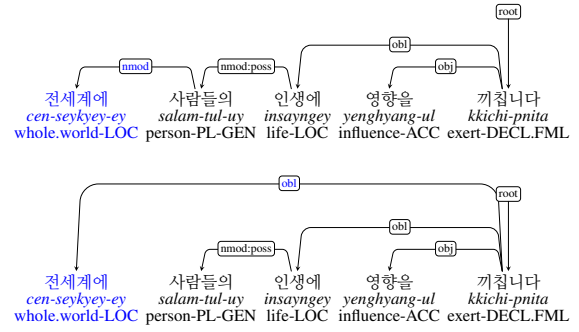

    \centering
    
    \resizebox{\linewidth}{!}{
        \begin{dependency}
            \begin{deptext}
            \textcolor{blue}{전세계에} \& 사람들의 \& 인생에 \& 영향을 \& 끼칩니다 \\
            \textcolor{blue}{\textit{cen-seykyey-ey}}  \& \textit{salam-tul-uy} \& \textit{insayngey} \& \textit{yenghyang-ul} \& \textit{kkichi-pnita}\\
            \textcolor{blue}{whole.world-LOC}  \& person-PL-GEN \& life-LOC \& influence-ACC \& exert-DECL.FML \\
            \end{deptext}
            \depedge{2}{1}{\textcolor{blue}{nmod}}
            \depedge{3}{2}{nmod:poss}
            \depedge{5}{3}{obl}
            \depedge{5}{4}{obj}
            \deproot{5}{root}
        \end{dependency}
    }
    
    \vspace{0.8em} 
    
    \resizebox{\linewidth}{!}{
        \begin{dependency}
            \begin{deptext}
            \textcolor{blue}{전세계에} \& 사람들의 \& 인생에 \& 영향을 \& 끼칩니다 \\
            \textcolor{blue}{\textit{cen-seykyey-ey}}  \& \textit{salam-tul-uy} \& \textit{insayngey} \& \textit{yenghyang-ul} \& \textit{kkichi-pnita}\\
            \textcolor{blue}{whole.world-LOC}  \& person-PL-GEN \& life-LOC \& influence-ACC \& exert-DECL.FML \\
            \end{deptext}
            \depedge{5}{1}{\textcolor{blue}{obl}}
            \depedge{3}{2}{nmod:poss}
            \depedge{5}{3}{obl}
            \depedge{5}{4}{obj}
            \deproot{5}{root}
        \end{dependency}
    }
    \caption{Modifier attachment ambiguity in a locative phrase. The locative 전세계에 (‘whole.world-LOC’) attaches either to 사람들 (‘people’) as \texttt{nmod} or to 끼칩니다 (‘exert’) as \texttt{obl}. Despite the locative marker (-에), the first analysis yields a possessive-like reading (‘people of the whole world’). \footnotesize{(Translated as Top: ‘[It] affects the lives of people in the whole world.’ Bottom: ‘In the whole world, [it] affects people’s lives.’)}}
\label{fig:8}    
\end{figure}

Finally, modifier attachment ambiguity (e.g., \texttt{amod–acl}, \texttt{nmod–obl}) reflects uncertainty in hierarchical scope, particularly when linear proximity does not clearly determine attachment. As illustrated in Figure~\ref{fig:7}, an adnominal form can be analyzed either as a lexical adjectival modifier (\texttt{amod}) or as a reduced relative clause (\texttt{acl}). Similarly, Figure~\ref{fig:8} shows that a locative phrase can attach either to a noun phrase (\texttt{nmod}) or to the predicate as a clausal oblique (\texttt{obl}), depending on its interpreted scope. Such modifier-attachment ambiguity is likewise context-dependent, requiring broader interpretive information beyond local morphosyntactic cues.

\subsubsection{Morphological-level disagreements}

Table~\ref{tab:7} presents the twenty most frequent morpheme-level XPOS mismatches. Similar to dependency relations, these disagreements form recurrent patterns rather than occurring randomly. First, many mismatches involved case particles (e.g., \texttt{JKS}, \texttt{JKB}, \texttt{JKO}, \texttt{JKC}, \texttt{JX}, \texttt{JKG}). These contrasts often reflect functional ambiguity, particularly in distinguishing structural case markers from auxiliary or semantic/discourse particles. Such morphological ambiguity closely parallels the dependency-level ambiguities observed in grammatical-relation identification and discourse structure.

\begin{table}[!]
\centering
\resizebox{0.9\linewidth}{!}{
\begin{tabular}{r l l r}
\hline
\textbf{Rank} & \textbf{Stanza} & \textbf{Trankit} & \textbf{Count} \\
\hline
1  & \texttt{NNG+\textbf{JKS}} & \texttt{NNG+\textbf{JKC}} & 54 \\
2  & \texttt{NNG+\textbf{JC}} & \texttt{NNG+\textbf{JKB}} & 48 \\
3  & \texttt{\textbf{NNG}+XSA+ETM} & \texttt{\textbf{XR}+XSA+ETM} & 40 \\
4  & \texttt{\textbf{VV}+ETM} & \texttt{\textbf{NV}+ETM} & 28 \\
5  & \texttt{\textbf{NNG}+VCP+EF} & \texttt{\textbf{NNB}+VCP+EF} & 27 \\
6  & \texttt{\textbf{VV}+EC} & \texttt{\textbf{NV}+EC} & 25 \\
7  & \texttt{--} & \texttt{JX} & 23 \\
8  & \texttt{MAG} & \texttt{NNG} & 23 \\
9  & \texttt{\textbf{VV}+ETM} & \texttt{\textbf{VX}+ETM} & 22 \\
10 & \texttt{\textbf{NNG}+JKB} & \texttt{\textbf{NF}+JKB} & 22 \\
11 & \texttt{\textbf{VX}+EC} & \texttt{\textbf{VV}+EC} & 21 \\
12 & \texttt{\textbf{NNG}+XSA+EC} & \texttt{\textbf{XR}+XSA+EC} & 20 \\
13 & \texttt{\textbf{NF}+JKO} & \texttt{\textbf{NNG}+JKO} & 20 \\
14 & \texttt{NNG} & \texttt{MAG} & 19 \\
15 & \texttt{NNG+NNG+NNG+\textbf{JKG}} & \texttt{NNG+NNG+NNG+\textbf{JX}} & 18 \\
16 & \texttt{\textbf{NNG}+JKO} & \texttt{\textbf{NF}+JKO} & 18 \\
17 & \texttt{VV+ETM+NNB} & \texttt{VV+ETM+NNB+\textbf{JX}} & 18 \\
18 & \texttt{VV+EC+VX} & \texttt{VV+EC+VX+\textbf{EC}} & 17 \\
19 & \texttt{\textbf{VA}+EC} & \texttt{\textbf{VV}+EC} & 17 \\
20 & \texttt{VV+EC+\textbf{VX}} & \texttt{VV+EC} & 17 \\
\hline
\end{tabular}
}
\caption{Twenty most frequent XPOS disagreement pairs between Stanza and Trankit. Counts indicate the number of tokens assigned different morpheme-level POS analyses.}
\label{tab:7}
\end{table}

Second, high-frequency mismatches often arose from differences in lexical decomposition and root identification (e.g., \texttt{NNG} vs. \texttt{XR}; \texttt{NNG} vs. \texttt{NF}; \texttt{VV} vs. \texttt{NV}). Two issues are implicated. First, the distinction between common nouns (\texttt{NNG}) and lexical roots (\texttt{XR}) is not always clear-cut in Korean, as root classification can be inherently ambiguous. Second, this ambiguity is further amplified in learner language, where non-canonical forms are frequent. In our annotation scheme, tags such as \texttt{NF}, \texttt{NV}, and \texttt{NA} mark ill-formed or irregular forms. While parsers may recover intended lexical items for recurring spelling errors through dictionary-based matching, novel or idiosyncratic errors often lack lexical support, resulting in divergent analyses. These patterns underscore the need for more systematic approaches to learner-specific morphological variation.

Finally, some disagreements involved segmentation differences, including the insertion or omission of functional morphemes (e.g., additional \texttt{JX} or \texttt{EC}). These cases reflect variation in morphological parsing strategies rather than simple tagging errors.

\section{Conclusion}

The purpose of this study was to evaluate a simplified HITL workflow for L2-Korean UD annotation. The findings provide three main implications, which may be relevant for researchers working on morphosyntactic annotation in learner corpora.

First, across all annotation features, the two independently domain-adapted parsers agreed on 82\% of token-level decisions. Within these consensus cases, human annotators also agreed on 93\% of instances. This strong alignment suggests that parser agreement reliably predicts correspondence with independent human judgments. For scalable L2 annotation, parser consensus may therefore serve as an effective filtering mechanism, substantially reducing the need for exhaustive manual verification. In addition, as noted by one reviewer, this binary agreement approach could be extended within an ensemble framework (e.g., \citealp{surdeanu2010ensemble}), which may enable more robust consensus estimation.

Second, despite high overall agreement, 31\% of tokens required human review in at least one feature, and 8\% required adjudication after initial correction. These findings suggest that morphosyntactic disagreement operates at multiple levels. Some cases are readily resolved through annotator agreement and are amenable to iterative model refinement, whereas others reflect deeper representational challenges in assigning L2-Korean forms to discrete morphosyntactic categories.

Third, parser disagreements clustered in linguistically predictable domains rather than occurring randomly. Our analysis showed that many involved argument-role distinctions and complement structures headed by complementizers, suggesting that targeted sampling and focused retraining could improve performance. In contrast, clause-boundary and modifier-attachment ambiguities were often context-dependent, indicating that some disagreements cannot be resolved through local morphosyntactic cues alone and may require broader contextual modeling. At the morphological level, frequent mismatches involved root identification, learner-specific spelling variation, and segmentation differences. These patterns highlight the need for systematic strategies to handle learner-generated forms, particularly when such non-standard forms are not attested in the training data or lexical resources.

In conclusion, we examined whether parser agreement can serve as a principled triaging mechanism in L2 annotation. The results point to its potential, while also highlighting the multi-level nature of morphosyntactic disagreement. Distinguishing between tractable modeling limitations and deeper representational ambiguities remains important for achieving efficient yet reliable analysis of learner language.

\section*{Limitations}

First, the dataset consists exclusively of argumentative writing by adult L2-Korean learners. Because parser performance may vary by proficiency, age, genre, and language background, the generalizability of these findings is limited.

Second, although we briefly noted issues related to spelling-error tags, this study did not systematically examine learner-specific morphological variation. Developing principled approaches to modeling such variation is therefore an important direction for future research.

Third, the proposed workflow reflects a simplified HITL design rather than a fully integrated, interactive system. For instance, it did not incorporate dynamic confidence estimation, active learning, or real-time model updating. Research in this line would benefit from incorporating these aspects into the HITL design.

\section*{Acknowledgments}
This study was supported by the 2024 Korean Studies Grant Program of the Academy of Korean Studies (AKS-2024-R-012). The authors gratefully acknowledge Youkyung Sung and Chanyoung Lee for their contributions to manual annotation, and Jeong Eun Shin for providing the data.


\newpage 

\bibliography{custom}

\appendix

\section{Gloss abbreviations}
\label{app:gloss}
Abbreviations used in interlinear glosses follow standard Leipzig conventions. 

\begin{table}[h]
\begin{tabular}{ll}
\hline
\textbf{Abbreviation} & \textbf{Meaning} \\
\hline
ACC   & Accusative \\
ADN   & Adnominal \\
ADV   & Adverbial \\
CONN  & Connective ending \\
COP   & Copula \\
DECL  & Declarative \\
FML   & Formal speech level \\
LOC   & Locative \\
NOM   & Nominative \\
PASS  & Passive \\
PL    & Plural \\
TOP   & Topic \\
\hline
\end{tabular}
\end{table}

\end{document}